\documentclass[review]{elsarticle}

\usepackage{lineno,hyperref}
\usepackage{hyperref}
\usepackage[cmex10]{amsmath}
\usepackage{color}
\usepackage{graphicx}
\usepackage{caption}
\usepackage{subcaption}
\usepackage{amssymb}
\usepackage{mathabx}
\usepackage{algorithm}
\usepackage{algorithmic}
\usepackage{balance}

\newcommand\smvee{\raise0.05ex\hbox{$\scriptstyle{\vee}$}}
\newcommand\smwedge{\raise0.05ex\hbox{$\scriptstyle{\wedge}$}}

\modulolinenumbers[5]

\journal{Pattern Recognition}

\begin{document}

\begin{frontmatter}

\title{Viewpoint Invariant Action Recognition\\ using RGB-D Videos}



\author[mymainaddress]{Jian Liu}
\author[mymainaddress]{Naveed Akhtar\corref{mycorrespondingauthor}}
\cortext[mycorrespondingauthor]{Corresponding author}
\ead{naveed.akhtar@research.uwa.edu.au}

\author[mymainaddress]{Ajmal Mian}

\address[mymainaddress]{School of Computer Science and Software Engineering, The University of Western Australia, 35 Stirling Highway, Crawley, 6009 WA, Australia.}

\begin{abstract}
In video-based action recognition, viewpoint variations often pose major challenges because the same actions can appear different from different views.
We use the complementary RGB and Depth information from the RGB-D cameras to address this problem.
The proposed technique capitalizes on the  spatio-temporal information available in the two data streams to the extract action features that are largely insensitive to the viewpoint variations.
We use the RGB data to compute dense trajectories that are translated to viewpoint insensitive deep features under a non-linear knowledge transfer model.
Similarly, the Depth stream is used to extract CNN-based view invariant features on which Fourier Temporal Pyramid is computed to incorporate the temporal information. 
The heterogeneous features from the two streams are combined and used as a dictionary to predict the label of the test samples.
To that end, we propose a sparse-dense collaborative representation classification scheme that strikes a balance between the discriminative abilities of the dense and the sparse representations of the samples over the extracted heterogeneous dictionary.   
To establish the effectiveness of our approach, we benchmark it on three standard datasets and compare its performance with twelve existing methods. 
Experiments show that the proposed approach achieves up to  7.7\%  improvement in the accuracy over its nearest competitor.

\end{abstract}

\begin{keyword}
Action recognition, viewpoint invariance, RGB-D videos.
\end{keyword}

\end{frontmatter}


\section{Introduction}
\label{sec:Int}
Video-based human action recognition is a challenging problem because of the large intra-action variations resulting from different illumination conditions, scales of videos, object textures and different scene backgrounds.
In the real-world scenarios, the task becomes even more arduous due to the camera viewpoint variations. 
Recent advances in the Depth video technology has resulted in mitigating  the problems arising from texture and illumination variations in the scenes because the Depth data is largely insensitive to such diversities. 
Nevertheless, the Depth videos still remain easily influenced by the camera viewpoint variations.
To incorporate viewpoint invariance in video-based human action recognition, we propose to fuse view-invariant deep features from RGB and Depth data streams and use the resulting heterogeneous  features for action recognition.


It is a common practice in the RGB video based action recognition~\cite{nCTE,Dense_traj_1,Dense_traj_2,Dense_traj_3} to extract the optical flow and dense trajectory features from the data and use them for predicting the action labels.
RGB videos generally  contain high-fidelity  appearance information about the scenes and the extracted dense trajectories provide important temporal cues that are useful in classifying an action. 
However, dense trajectories are sensitive to the camera viewpoints~\cite{HPM+TM}.
Moreover, such features  do not contain any explicit information regarding the human poses in the videos.
It is easy to imagine that human pose information can be useful in a more accurate action recognition in realistic conditions.


Rahmani and Mian~\cite{NKTM} recently proposed a Non-linear Knowledge Transfer Model~(NKTM) for cross-view action recognition. 
Their model maps dense trajectory features of RGB videos to a single canonical viewpoint, that makes their approach less sensitive to the viewpoint variations.
Nevertheless, their technique is  unable to take full advantage of the human pose information because it is harder to extract such details using only the RGB data.
Comparatively, the Depth data  is much more suitable for extracting the human poses.
Thus, their approach becomes a sub-optimal choice for the cases where the Depth data is also easily available. 
To take advantage of the human pose information in videos, Rahmani and Mian~\cite{HPM+TM} separately proposed a Human Pose Model (HPM), that extracts view-invariant features using the Depth stream only. 
Accurate action recognition using the HPM features~\cite{HPM+TM} substantiates the usefulness of the pose information in action recognition.


With the recent easy availability of the RGB-D sensors (e.g.~Kinect), it is intuitive to jointly exploit both RGB and  Depth data streams for  action recognition. 
Existing approaches~\cite{Chaaraoui_2013_ICCVW,Hu_2015_CVPR,Liu_2013_IJCAI,Shahroudy_2014_ISCCSP,Koppula_2013_ICML,Wei_2013_ICCV} also report reasonable performance gain with the joint usage of the two data streams.
However, none of these approaches are tailored to incorporate viewpoint invariance in action recognition. 
Whereas RGB and Depth data streams of RGB-D videos  naturally complement each other, specialized processing of each stream is imperative for successful  viewpoint invariant action recognition. 
Moreover, to take full advantage of both data streams it is important to carefully combine them to amplify the view-invariant character of the resulting data.

\begin{figure*}[t]
\centering
\includegraphics[width=1\textwidth]{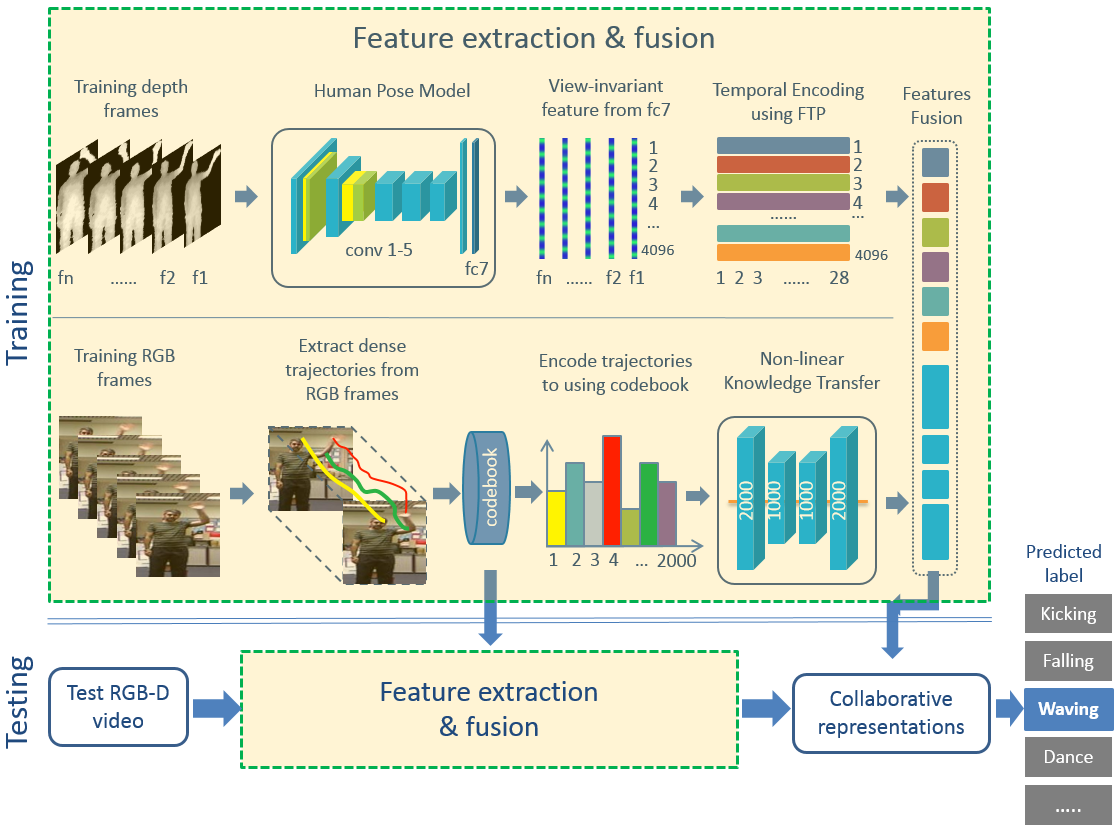}
\caption{Schematics of the proposed approach: During training, the Depth stream extracts CNN features using the Human Pose Model~\cite{HPM+TM} and performs temporal endcoding using the FTP. The RGB stream is used to extract  dense trajectories and a deep network, implementing non-linear knowledge transfer, is employed to extract the view invariant features. The resulting heterogeneous features are fused together and used as a set of basis vectors in collaborative representation based classification of the test samples.}
\label{fig:schema}
\end{figure*}

In this work, we propose a view-invariant action recognition approach, illustrated in Fig.~\ref{fig:schema}, that accounts for the aforementioned observations.
The proposed approach, that builds on our recent work~\cite{DICTA}, simultaneously extracts view-invariant features from the RGB and the Depth data streams and meticulously combines these heterogeneous features for better classification. 
For the Depth features, it exploits the  CNN-based Human Pose Model~\cite{HPM+TM} and performs temporal encoding of the extracted CNN features using the Fourier Temporal Pyramid.
For the RGB stream, we first extract dense trajectory features and encode them with a representative code book. The resulting codes are passed through another deep network model to extract the desired view invariant features. 
The network is implemented using the Caffe library~\cite{jia2014caffe} and it has been learned from scratch for extracting the view-invariant  features.

Our approach carefully combines the RGB and the Depth features and uses the joint features as a dictionary to classify the test samples.
To that end, we propose to take advantage of both dense and sparse representations of a test feature over the extracted heterogeneous dictionary. We propose a classification scheme that balances between the discriminative abilities of dense and sparse representations to achieve the optimal performance under the collaborative representation based classification framework~\cite{CRC}.
We evaluate our approach on three standard multi-view action recognition  datasets and compare its performance with twelve existing techniques. 
Results of the performed experiments clearly establish the effectiveness of the proposed approach. 

The remaining article  is organized as follows.
In Section~\ref{sec:RW}, we discuss the related existing literature in RGB, Depth and RGB-D action recognition. 
We explain the proposed technique in Setion~\ref{sec:PA} and report detailed experimental results in Section~\ref{sec:Exp}.
The paper is concluded in Section~\ref{sec:Conc}.

\section{Related work}
\label{sec:RW}

RGB and Depth (D) are the two major video data formats used in human action recognition. In this section, we discuss the related work relevant  to each of these formats as well as the combined RGB-D format. 

\subsection{RGB Video Human Action Recognition}
In  RGB video based action recognition, few exiting approaches~ \cite{rw_3,rw_4,rw_5} directly use  geometric transformations to incorporate the much needed viewpoint invariance.
However, to achieve the desired performance level, it is   critical for these methods to accurately estimate the skeleton joints. 
In practical conditions, it is often challenging to achieve high level of accuracy in skeleton joint estimation, which makes these methods less appealing for the practical purpose.
Another stream of techniques~\cite{Hankelets,rw_7,rw_8,rw_9} exploits spatio-temporal  features in the RGB videos to incorporate the viewpoint invariance. 
Nevertheless, the action recognition performance of these approaches is generally limited by the structure of the extracted features~\cite{HPM+TM}. 


Another popular framework in RGB video based view-invariant action recognition is to find a latent space where the features are insensitive to viewpoint variations  and classify the actions employing  that latent space~\cite{rw_10,rw_11,rw_12,DVV,rw_14,CVP,rw_16}.
A combination of hand-crafted features and deep-learned features was also proposed by Wang et al.~\cite{Wang_2015_CVPR} for the RGB action recognition. In their approach, trajectory pooling was used for one stream of the data and deep learning framework was used for the other. The two feature streams were combined to form trajectory-pooled deep-convolutional descriptors. Nevertheless, only RGB data is used in their approach and the problem of viewpoint variance in action recognition is not directly  addressed. 


\subsection{Depth Video Human Action Recognition}
With the easy availability of the Depth data through the Microsoft Kinect sensor, the Depth video based action recognition became much popular in the last decade.
In \cite{rw_19} and \cite{rw_17}, 3D data points are used at silhouettes of the Depth images and 3D joint positions to extract features for  action recognition.
A binary range-sample feature was proposed for the Depth videos by Lu $et$ $al$ \cite{Lu_2014_CVPR}, that demonstrated significant improvement in achieving viewpoint invariance. 
Rahmani $et$ $al$~\cite{HOPC} proposed the Histogram of Oriented Principal Components (HOPC) to detect interest points in Depth videos and extracting their spatio-temporal descriptors. HOPC extracts local features in an object-centered local coordinate basis, thereby making them viewpoint invariant. Nevertheless, these features must be extracted at a lagre number of interest points that makes the overall approach computationally expansive.
In another work, Yang et al.~\cite{SNV} clustered hypersurface normals in the Depth sequences to characterize the local motions and the shape information. An adaptive spatio-temporal pyramid is used in their approach to divide the Depth data into a set of space-time grids and low-level polynomials are aggregated to form a Super Normal Vector (SNV). This vector is eventually employed in action recognition.


\subsection{RGB-D Video Human Action Recognition}
Most of the Depth sensors also provide simultaneous RGB videos. This fact has lead to a significant interest of the scientific community to jointly exploit the two data streams for various tasks, including action recognition~\cite{Chaaraoui_2013_ICCVW,Shahroudy_2014_ISCCSP,Koppula_2013_ICML,Wei_2013_ICCV,Liu_2013_IJCAI,Hu_2015_CVPR,Kong_2015_CVPR}.
For instance, a restricted graph-based genetic programming approach is proposed by Liu and Shao \cite{Liu_2013_IJCAI} for the fusion of the Depth and the RGB data streams for improved RGB-D data based classification. 
In another approach, Hu et al.~\cite{Hu_2015_CVPR} proposed to learn heterogeneous features for the RGB-D video based action recognition.
They proposed a joint heterogeneous features learning model (JOULE) to take advantage of both shared and action-specific components in the RGB-D videos.
Kong and Fu \cite{Kong_2015_CVPR} also projected and compressed both Depth and RGB features to a shared feature space, in which the decision boundaries are learned for the classification purpose.

Whereas one of the major advantages of the Depth videos is in the easy availability of the information useful for the problem of viewpoint invariant action recognition, none of the aforementioned approaches directly address this problem. 
Moreover, the Depth and the RGB video frames are mainly combined in those approaches by either projecting them to a common feature space \cite{Hu_2015_CVPR, Kong_2015_CVPR} or by using the same  filtering-pooling operation for both modalities \cite{Liu_2013_IJCAI}.
We empirically verified that on the used action recognition datasets that involve multiple camera views, these techniques achieve no more than 4\% improvement (RGB-D combined) over the single (RGB or Depth) modality.
On the other hand, the technique proposed in this work achieves up to { 7.7\%} average improvement over the single modality while  dealing with the viewpoint variations. 
The strength of our approach resides in processing the RGB and the Depth streams individually to fully capitalize on their individual characteristics and then fusing the two modalities at a latter stage of the pipeline. This strategy has proven much more beneficial than combining the two data streams  earlier in data processing.

\section{Proposed approach}
\label{sec:PA}
The proposed RGB-D based human action recognition approach is illustrated in Fig.~\ref{fig:schema}.
We use a Deep Learning based Human Pose Model \cite{HPM_code} to extract features from the Depth data stream.
These features are post processed by computing Fourier Temporal Pyramids. 
For the RGB stream, we extract dense trajectory features and encode them over a codebook. The resulting codes  are passed trough a network implementing a non-linear knowledge transfer model. 
The two types of features are combined and used as a dictionary in our approach. 
To classify, we first extract the joint features of the test samples and then collaboratively represent them over the extracted dictionary. 
These representations are used in predicting the class labels. 

\subsection{Depth Feature Extraction}
\label{sec:EDF}

We extract CNN-based features from the Depth data stream using a pre-trained Human Pose Model (HPM)~\cite{HPM+TM}. 
HPM uses an architecture structure similar to the AlexNet~\cite{AlexNet} but it has been trained using multi-viewpoint action data that is generated synthetically by fitting human models to the CMU motion caption data~\cite{CMU}.
Using data from 180 different viewpoints for each action makes the network training largely insensitive to the view variations in the actions. 
HPM essentially extracts features using Depth \emph{images} with static human poses.
Thus, to incorporate the temporal dimension we use the Fourier Temporal Pyramid (FTP)~\cite{rw_21} on the HPM features. 

Let us denote the $t^{\text{th}}$ frame of the $i^{\text{th}}$ Depth video by ${\bf V}_t^i$, where $t \in \{1,2,...,f\}$, such that $f$ represents the total number of frames. 
We first crop a Depth frame ${\bf V}^i \in \{ {\bf V}_1^i, {\bf V}_2^i,...{\bf V}_f^i\}$ to resize it to a $227\times227$ image in order to match the input dimensions of the HPM.
These frames are passed through the HPM and the $fc7$ layer output activation vector ${\bf a}_t^i \in \mathbb R^{4096}$ is used as the feature for the Depth frame ${\bf V}_t^i$. 
The features from the $i^{\text{th}}$ video are combined into a matrix ${\bf A}^i = [{\bf a}_1^i, {\bf a}_t^i,..., {\bf a}_f^i] \in \mathbb R^{4096 \times f}$ and the Fourier Temporal Pyramid of the matrix is computed for the temporal encoding.
More specifically, the  computed pyramid has 3 levels, and by dividing the feature set ${\bf A}^i$ in half at each level, we get $1+2+4=7$ feature groups. Short Fourier Transform is applied to each feature group, and the first 4 low-frequency coefficients are concatenated (i.e. $4 \times 7 = 28$) to form a spatial-temporal descriptor matrix ${\bf S}^i \in \mathbb{R}^{4096 \times 28}$ for an  action. Finally, the matrix ${\bf S}^i$ is vectorized as ${\bf d}^i \in \mathbb{R}^{ 114688}$ to get the Depth feature for the $i$-th Depth video.

\subsection{RGB Feature Extraction} \label{RGB_feature_extraction}
For our approach, we employ another neural network to extract the RGB features. 
The network model is implemented using the Caffe library~\cite{jia2014caffe} and it has been trained from the scratch for the proposed approach.
Momentarily, we defer the discussion on the model details and describe the preprocessing of the data for its training.  
Similar to the HPM, we use the CMU motion capture data~\cite{CMU} for training this network.
The data pre-processing is conducted as follows. 
We first fit cylinders to the available skeleton data in order to approximate the human limbs and the torso.
The points on those cylinders are then rendered from eighteen different viewpoints using their orthographic projections onto the x-y plane.  
We extract the dense trajectory features~\cite{nCTE} from the resulting data. 
In order to do that, we follow Gupta et al.~\cite{nCTE} and fix the frame length to $L = 15$. 
This results in thirty-dimensional trajectory features. 
Notice that we again use the data generated from  multiple viewpoints, which is important to introduce viewpoint invariance in the RGB features, similar to the Depth features. 
Once the trajectory features are extracted, we perform clustering over these features using the K-Means algorithm, where we empirically choose $K = 2000$.  
We store the cluster centroids in $ {\bf C} \in \mathbb R^{30 \times 2000}$. 
The matrix ${\bf C}$ represents a codebook, containing the $2000$ most representative trajectories for the data.

\begin{figure}[t]
\centering
\includegraphics[width=\columnwidth]{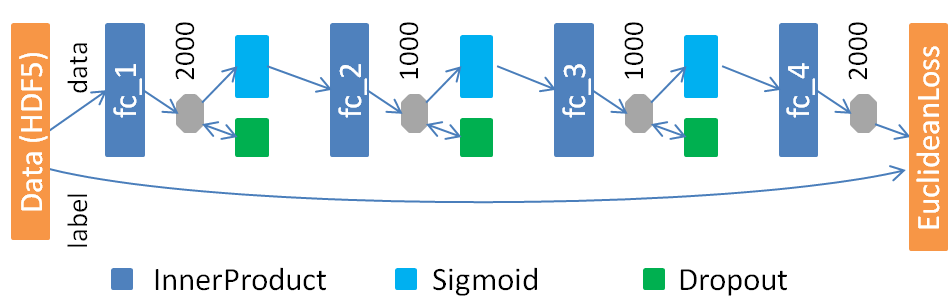}
\caption{Caffe implementation of the used network. Dropout and sigmoid activation are applied on the first 3 inner product layers.}
\label{fig:caffe_nktm_net}
\end{figure}

To train the network, a video is first encoded over the matrix ${\bf  C}$.
The coding is performed such that the resulting Bag of word  $\boldsymbol \xi \in \mathbb R^{2000}$  represents a histogram of video trajectories closest to each column of the matrix ${\bf  C}$ in the Euclidean space. 
Our neural network implements the non-linear knowledge transfer model~\cite{NKTM} that projects the vectors $\boldsymbol \xi$ of all the  viewpoints of an action to a single canonical (i.e. frontal) viewpoint.
The model essentially regresses for the problem of projecting multiple features to a single feature, thereby not requiring any explicit label during the training. 
The architecture of the network based on our Caffe implementation is shown in Fig.~\ref{fig:caffe_nktm_net}.
The network comprises four layers with drop-outs and sigmoid activations applied to the first three layers. 
The main intuition behind the chosen architecture is that since we must keep the regression flexible, in the sense that it is able to project different views to a single view, we first reduce the layer size to  drop redundant information in the features.
In the latter layer, the size is again increased to discriminate between the view specific high level details of the inputs.   

\begin{figure}[t]
\centering
\includegraphics[width=\columnwidth]{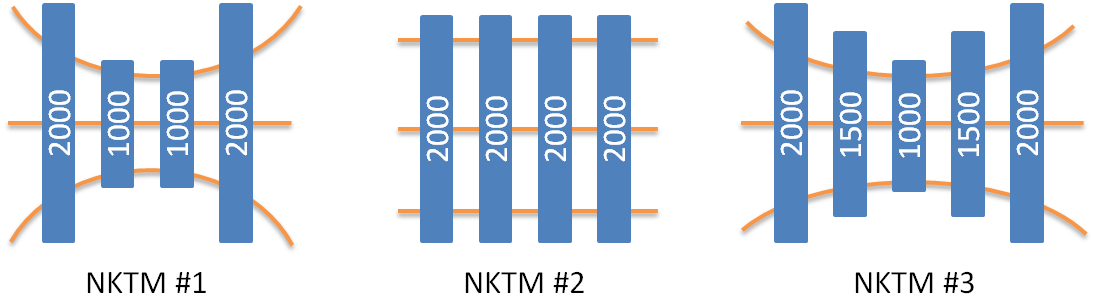}
\caption{Three network architectures trained using the Caffe library \cite{jia2014caffe}.}
\label{fig:caffe_nktm_modes}
\end{figure}

In light of the original proposal of the non-linear knowledge transfer approach by Rahmani and Mian~\cite{NKTM}, we tested multiple architectural variations of the above mentioned network. 
In Fig~\ref{fig:caffe_nktm_modes}, we show the three variations for which we also present the  experimental results in Section~\ref{sec:Exp} to illustrate the sensitivity of our approach to the architecture of the network. 
The chosen network is labeled NKTM \#1 in Fig.~\ref{fig:caffe_nktm_modes} (for the Non-linear Knowledge Transfer Model).
For training, we initialize the network with the weight filler ``xavier'' and variance norm ``AVERAGE'', and configure the loss layer as the ``EuclideanLoss''. We use the initial learning rate of 0.001 for all the layers and decrease it by a factor of 10 after every 1000 iterations. The weight decay of the network is configured to 0.0005. We train our network using the back-propagation algorithm with six thousand iterations.
These parameters are chosen empirically in our approach.

Once the network is trained using the CMU mocap data we use it to extract features of the action recognition training data.
For that, we first extract their dense trajectories following~\cite{Dense_traj_1,Dense_traj_2}. These trajectories are coded over the already learned codebook ${\bf C}$ and the resulting codes are passed through the trained network. In order to finally arrive at the view invariant features, we concatenate the outputs of each layer of the network.
It was empirically verified for the used network that its earlier layers are more informative for the small variations in the viewpoints whereas the latter layers are more informative for the larger viewpoint variations.
Therefore, we concatenate the  outputs from all the layers to construct the RGB feature vector  ${\bf r}\in \mathbb R^{6000}$.

\subsection{Feature Fusion}
The RGB and the Depth features are extracted from two different data streams in our approach.
However, they contain complementary information due to the following  reasons.
Firstly, the data streams originate from the same actions. Secondly, the dense trajectories model the temporal information from the RGB frames using only  the motion (without explicitly encoding the  poses), whereas the Depth features are extracted by modeling the human poses from the raw Depth images (and using the FTP). 
Thus, the latter is specifically meant to complements the former.
It is also worth noting that the dense trajectories are better extracted from the RGB frames due to the presence of texture.
On the other hand, human poses are more accurately computed using the Depth images because they contain the shape information.

\begin{figure}[t]
\centering
\includegraphics[width=3.3in]{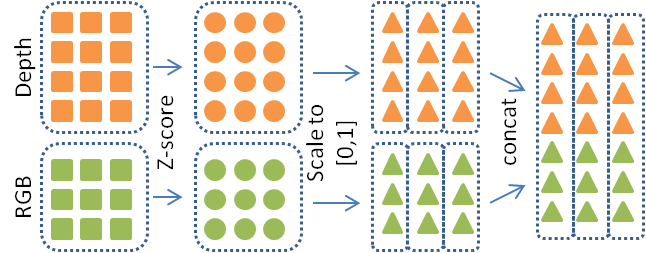}
\caption{Illustration of feature fusion process. First, the Z-scores of the Depth and the RGB features are computed. Then, sample-wise scaling is performed to restrict  features in range [0,1].Finally, a row-wise concatenation is performed.}
\label{fig:features_norm}
\end{figure}

To integrate the strengths of the RGB and the Depth features, we propose to fuse them into a joint feature. 
We use these heterogeneous joint  features as the basis vectors to collaboratively represent the test sample features and classify the resulting representations. 
The proposed feature fusion scheme works as follows.
Let ${\bf D}=[{\bf d}^1, {\bf d}^2, \dots, {\bf d}^n] \in \mathbb R^{114688 \times n}$ and ${\bf R}=[ {\bf r}^1, {\bf r}^2, \dots {\bf r}^n] \in \mathbb R^{6000 \times n}$  denote the feature sets obtained from the Depth and the RGB  streams respectively.
We first transform ${\bf D}$ and ${\bf R}$ into  ${\bf D}_z\in \mathbb R^{114688 \times n}$ and ${\bf R}_z \in \mathbb R^{6000 \times n}$, where the latter matrices are computed as the Z-scores of the formers. Then, the columns of ${\bf D}_z$ and ${\bf R}_z$ are rescaled to the range $[0,1]$.
Finally, the rescaled features are row-wise concatenated to form the  heterogeneous feature set ${\bf X} \in \mathbb{R}^{120688  \times n}$.
The procedure is also illustrated in Fig.~\ref{fig:features_norm}.

The feature fusion procedure is specifically designed keeping in view the requirements of the proposed approach.
In our framework, the obtained Depth and RGB features generally vary greatly in terms of scale and dimensionality.
However, we still aim at using the combined features as the joint basis vectors to represent the test sample features.
Therefore, we first transform the  features to a dimensionality independent measure of Z-score, which represents each coefficient of a feature vector in terms of the standard deviation of the vector components. 
A further rescaling is performed to remove the scale differences of the  extracted features.
Once the rescaled features are combined, they form an effective set of basis vectors for the  subspace in which the test sample features reside.

\subsection{Classification}
Our label prediction stage is inspired by the Collaborative Representation (CR) based classification framework~\cite{NA_PR}.
To predict the label of a test sample, we first extract its joint feature ${\bf y} \in \mathbb R^{114688}$ following the  feature extraction  procedure for the training data.
Then, we compute two separate collaborative representations of the test feature in terms of the training features by solving the following optimization problems:
\begin{align}
\boldsymbol{\widecheck\alpha} = \min_{\alpha} ||{\bf y} - {\bf X} \boldsymbol\alpha||_2, + \lambda ||\boldsymbol\alpha||_2 
\label{eq:opt1}
\end{align}
\begin{align}
\widehat{\boldsymbol\alpha} = \min_{\alpha} ||{\bf y} - {\bf X} \boldsymbol\alpha||_2,~s.t.~||\boldsymbol\alpha||_0 \leq k,
\label{eq:opt2}
\end{align}
where,  $||.||_p$ denotes the $\ell_p$-norm of a vector and $k$ is the sparsity threshold.

The vector $\boldsymbol{\widecheck\alpha} \in \mathbb R^n$ is computed by regularizing the $\ell_2$-norm of the coefficient vector, that allows most of the components of $\boldsymbol{\widecheck\alpha}$ to be non-zero in the representation.  
This makes $\boldsymbol{\widecheck\alpha}$ the \emph{dense} representation of ${\bf y}$ over the matrix ${\bf X}$.
On the other hand, the sparsity constraint in (\ref{eq:opt2}) forces most of the coefficients in the representation vector to become zero, thereby making  $\widehat{\boldsymbol\alpha} \in \mathbb R^n$ the \emph{sparse} representation of ${\bf y}$. 
Since the basis vectors in $\bf X$ have heterogeneous composition in our approach, CR-based classification framework can be expected to favor sparse representations in few instances and dense ones for the other.  
Therefore, we propose to combine the two representations in a convex manner and use this combination for predicting the class labels of the test samples. 
More precisely, we compute $
{\boldsymbol{\overset{\circ}\alpha}} = \lambda_1\widehat{\boldsymbol\alpha} + (1-\lambda_1)\boldsymbol{\widecheck\alpha}$, where $\lambda_1 \leftarrow [0,1]$ and later employ  $\boldsymbol{\overset{\circ}\alpha} \in \mathbb R^n$ to predict the class label. 
We note that this representation combination  strategy differs from \cite{NA_PR} that augments a dense collaborative representation with a sparse one by simply adding the two and normalizing the resulting vector.
Convex combination of the two types of representations ascertain global optimality of the solution with respect to the parameter $\lambda_1$. This guarantee is not provided by the data augmentation method proposed in \cite{NA_PR}.

In  CR-based classification, it is generally the case that the class label is predicted either by maximizing the reconstruction fidelity of the collaborative representation coefficients~\cite{CRC} or by using a multi-class classifier in conjunction with the computed representation~\cite{PAMI}, \cite{NA_CVPR2017}.
However, these methods are computationally expensive.
We recently showed~\cite{NA_PR} that collaborative representations can be used for predicting the class labels much more efficiently by integrating over their coefficients for each class and assigning the class label to the largest integrated value.
Here, we implement this strategy by multiplying $\boldsymbol{\overset{\circ}\alpha}$ with a binary matrix ${\bf B} \in \mathbb R^{C \times n}$, where $C$ denotes the total number of classes. 
This matrix is constructed by assigning $1$ to the $i^{\text {th}}$ row of its $j^{\text {th}}$ column if the $j^{\text {th}}$ column of $\bf X$ belongs to the $i^{\text {th}}$ class of the training features. All other coefficients of the matrix are kept $0$. With a matrix thus constructed, each coefficient of the vector ${\bf q} = {\bf B}\boldsymbol{\overset{\circ}\alpha} \in \mathbb R^C$ integrates the coefficients of  $\boldsymbol{\overset{\circ}\alpha}$ for a single class.
We maximize over the coefficients of $\bf q$ to predict the class label of the test sample.
The proposed classification procedure is also summarized as Algorithm~\ref{alg:2}.

\floatname{algorithm}{Algorithm}
\renewcommand{\algorithmicrequire}{\textbf{Initializaiton:}}
\renewcommand{\algorithmicensure}{\textbf{Main Iteration:}}
\floatname{algorithm}{Algorithm}
\renewcommand{\algorithmicrequire}{\textbf {Input:}}
\renewcommand{\algorithmicensure}{\textbf{Output:}}
\begin{algorithm}
\caption{Sparse-Dense CR-based Classification} 
\begin{algorithmic}[1]
\REQUIRE (a) Training features  ${\bf X}$, normalized in $\ell_2$-norm. (b) Test sample feature ${\bf y}$. (c) Regularization parameters~$\lambda, \lambda_1$. (d)~Sparsity threshold $k$. (e) Binary matrix ${\bf B}$.
\\
\STATE  \emph{Optimization:}\\
a)~Solve (\ref{eq:opt1}) as $
\boldsymbol{\widecheck\alpha} = {\bf P}{\bf y},
\label{eq:RLS}
$
where, ${\bf P} = ({\bf X}^{\text{T}} {\bf X} + \lambda{\bf I}_n)^{-1}{\bf X}^{\text{T}}$.\\
b)~Solve (\ref{eq:opt2}) using the Orthogonal Matching Pursuit algorithm~\cite{OMP}:
\STATE \emph{Convex combination:}  Compute 
$
{\boldsymbol{\overset{\circ}\alpha}} = \lambda_1\widehat{\boldsymbol\alpha} + (1-\lambda_1)\boldsymbol{\widecheck\alpha}
$.
\STATE \emph{Labeling:} label$({\bf y}) = \text{arg max}_{i}\{q_i\}$, where $q_i$ denotes the $i^{\text{th}}$ coefficient of ${\bf q} = {\bf B}\boldsymbol{\overset{\circ}\alpha}$. 
\ENSURE  label$({\bf y})$. 
\end{algorithmic}
\label{alg:2}
\end{algorithm}

\section{Experiments}
\label{sec:Exp}
The proposed approach has been evaluate  on three multiview RGB-D datasets: UWA 3D Multiview Activity II Dataset \cite{HOPC2016PAMI,HOPC},  Northwestern-UCLA Multiview Action 3D Dataset \cite{AOG} and the NTU RGB+D Human Activity Dataset \citep{shahroudy2016ntu}. 
We  compare the performance of the proposed approach with the existing state-of-the-art methods. 
The results of the existing methods are taken directly from the original papers where applicable, otherwise, these  results are taken from the best  reported results in the literature. 
For the proposed approach, we used $\lambda = 0.01$,  $\lambda_1 = 0.35$ and the sparsity threshold $k = 50$. These parameters were optimized empirically using cross-validation. 

\subsection{UWA3D-II Dataset}
\begin{figure}[t]
\centering
\includegraphics[width=0.9\textwidth]{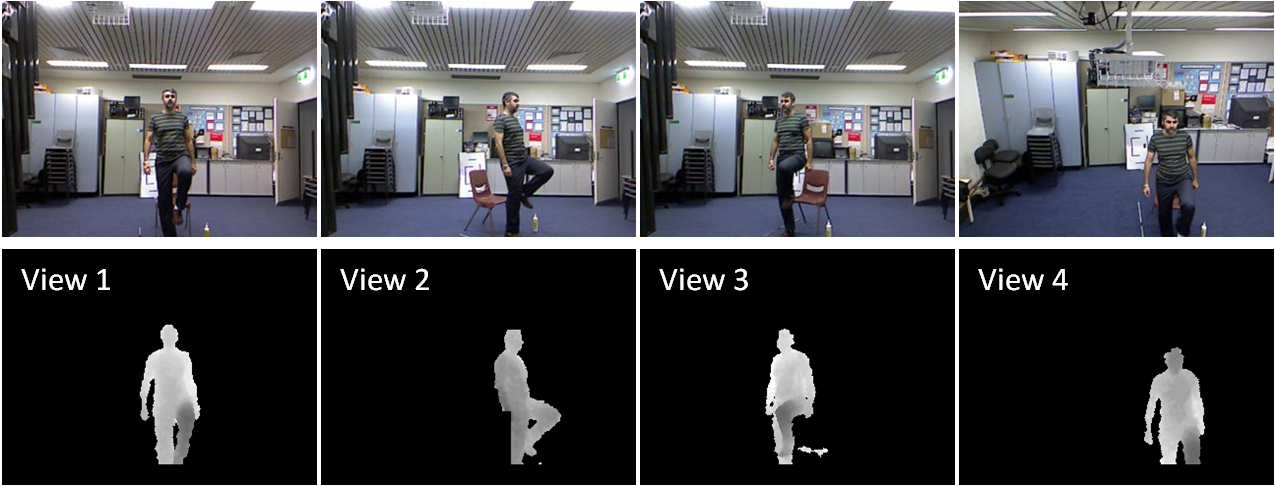}
\caption{RGB and Depth pairs from the UWA 3D Multiview Activity II Dataset \cite{HOPC}.}
\label{fig:uwa3d_sample}
\end{figure}

This dataset is composed of 30 human actions that have been performed by 10 subjects, where each action is recorded from 4 different viewpoints. 
These actions include: (1) one hand waving, (2) one hand punching, (3) two hands waving, (4)~two hands punching, (5) sitting down, (6) standing up, (7) vibrating, (8) falling down, (9) holding chest, (10) holding head, (11) holding back, (12) walking, (13) irregular walking, (14) lying down, (15) turning around, (16) drinking, (17) phone answering, (18) bending, (19) jumping jack, (20) running, (21) picking up, (22) putting down, (23) kicking, (24) jumping, (25) dancing, (26) moping floor, (27) sneezing, (28) sitting down (chair), (29) squatting, and (30) coughing. The four viewpoints are: (a) front, (b) left, (c) right, (d) top.
The dataset is challenging because its action classes are reasonably large and they additionally containing  subject scale and viewpoint variations. Furthermore, human-object interaction and self-occlusion in few videos makes action recognition on this dataset even more challenging. 
Figure \ref{fig:uwa3d_sample} shows a representative example  of RGB and Depth pair for one of the actions from different viewpoints.

For the evaluation, we follow Wang et al.~\cite{AOG} and use videos from two views for training and the remaining views for testing, creating 12 different view combinations. 
We first compare the results of the proposed approach employing three variants of the  non-linear knowledge transfer model network  architectures (see Fig ~\ref{fig:caffe_nktm_modes}). Note that, the original knowledge transfer model~\cite{NKTM} only deals with the RGB frames. 
We implement the network variants using the Caffe library and train our models from the scratch.
Table~\ref{tab:uwa_comp_nktm} summarizes the results of this comparison.
Based on these results we chose Network~\#1 for our approach.
Henceforth, unless otherwise mentioned, the results of the proposed approach are based on this architecture.


\begin{table*}[t]
\centering
\caption{Action recognition accuracies (\%) on UWA 3D-II dataset using variants of NKTM architectures shown in Fig~\ref{fig:caffe_nktm_modes}. The architectures are implemented using Caffe library and trained from the scratch.} 
\tabcolsep=0.076cm
{\footnotesize
\begin{tabular}{|l|c|c|c|c|c|c|c|c|c|c|c|c|c|}
\hline
Training view & \multicolumn{ 2}{c|}{V1 \& V2} & \multicolumn{ 2}{c|}{V1 \& V3} & \multicolumn{ 2}{c|}{V1 \& V4} & \multicolumn{ 2}{c|}{V2 \& V3} & \multicolumn{ 2}{c|}{V2 \& V4} & \multicolumn{ 2}{c|}{V3 \& V4} & \multicolumn{ 1}{c|}{Mean} \\ \cline{ 1- 13}
Testing view & V3 & V4 & V2 & V4 & V2 & V3 & V1 & V4 & V1 & V3 & V1 & V2 & \multicolumn{ 1}{c|}{} \\ \hline
\multicolumn{ 14}{|c|}{\textbf{Input: RGB + Depth images}} \\ \hline

NKTM \#1 & {\bf 86.9} & {\bf 89.8} & 81.9 &  89.5 & {\bf 76.7} & {\bf 83.6} & {\bf 83.6} & {\bf 79.0} & 89.6 & {\bf 82.1} & {89.2} & {\bf 83.8} & {\bf 84.6} \\ \hline

NKTM \#2 & 85.1 & 88.8 & {\bf 82.0} & 87.6 & 76.3 & 82.1 & 83.3 & {78.3} & {\bf 90.3} & 81.0 & 89.2 & 79.3 & 83.6 \\ \hline

NKTM \#3 & 85.1 & 88.4 & 80.8 & {\bf 89.9} & 74.4 & 82.1 & {83.6} & 76.4 & 90.0 & {81.3} & {\bf 90.0} & {82.3} & 83.7 \\ \hline

\end{tabular}}
\label{tab:uwa_comp_nktm}
\end{table*}
 
\begin{figure}[t]
\centering
\includegraphics[width=0.6\textwidth]{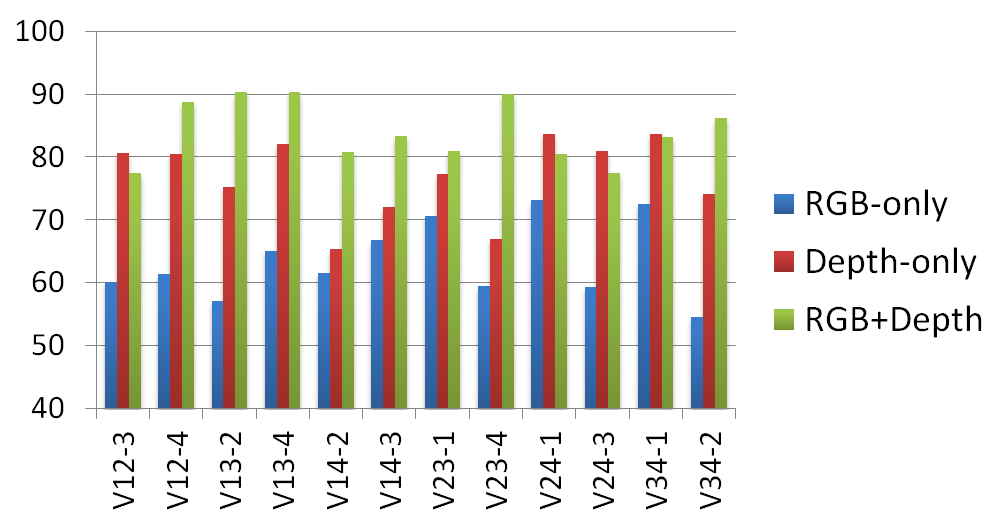}
\caption{Comparison of the proposed method with the best performing RGB-only and Depth-only methods on the UWA 3D-II dataset. V12-3 indicates that view 1 and 2 are used for training while view 3 is used for testing.}
\label{fig:uwa_comp_methods}
\end{figure}

Table~\ref{tab:uwa_comp_1}, summarizes the quantitative comparison of the proposed  approach with the existing methods.  
In the table, the proposed method achieves 84.6\% average recognition accuracy, which is 7.7\% higher than its nearest competitor HPM+TM \cite{HPM+TM}. 
In Fig.~\ref{fig:uwa_comp_methods} we compare our approach to the the best performing RGB-only method~\cite{NKTM} and the best performing Depth-only method~\cite{HPM+TM}.
Note that, for all of the train-test view combinations, the proposed method provides a  significant improvement in the recognition accuracy.
The maximum reduction in the error rate, i.e. 36.4\%, is achieved when view 2 and 3 are used for training and view 4 is used for testing. 
Based on these results, we can argue that our method effectively integrates the advantages of RGB and Depth video streams to enhance the recognition accuracy, especially for the large viewpoint variations.

\begin{table}[t]
\centering
\caption{Comparison of action recognition accuracy (\%) on the UWA3D-II dataset. Each time two views are used for training and the remaining two views are individually used for testing.}
\vspace{1mm}
\tabcolsep=0.07cm
{\footnotesize
\begin{tabular}{|l|c|c|c|c|c|c|c|c|c|c|c|c|c|}
\hline
Training views & \multicolumn{ 2}{c|}{V1 \& V2} & \multicolumn{ 2}{c|}{V1 \& V3} & \multicolumn{ 2}{c|}{V1 \& V4} & \multicolumn{ 2}{c|}{V2 \& V3} & \multicolumn{ 2}{c|}{V2 \& V4} & \multicolumn{ 2}{c|}{V3 \& V4} & \multicolumn{ 1}{c|}{Mean} \\ \cline{ 1- 13}
Testing view & V3 & V4 & V2 & V4 & V2 & V3 & V1 & V4 & V1 & V3 & V1 & V2 & \multicolumn{ 1}{c|}{} \\ \hline
\multicolumn{ 14}{|c|}{\textbf{Input: RGB images}} \\ \hline
AOG \cite{AOG} & 47.3 & 39.7 & 43.0 & 30.5 & 35.0 & 42.2 & 50.7 & 28.6 & 51.0 & 43.2 & 51.6 & 44.2 & 42.3 \\ \hline
Action Tube \cite{Action_Tube} & 49.1 & 18.2 & 39.6 & 17.8 & 35.1 & 39.0 & 52.0 & 15.2 & 47.2 & 44.6 & 49.1 & 36.9 & 37.0 \\ \hline
LRCN \cite{LRCN} & 53.9 & 20.6 & 43.6 & 18.6 & 37.2 & 43.6 & 56.0 & 20.0 & 50.5 & 44.8 & 53.3 & 41.6 & 40.3 \\ \hline
NKTM \cite{NKTM} & 60.1 & 61.3 & 57.1 & 65.1 & 61.6 & 66.8 & 70.6 & 59.5 & 73.2 & 59.3 & 72.5 & 54.5 & 63.5 \\ \hline
\multicolumn{ 14}{|c|}{\textbf{Input: Depth images}} \\ \hline
DVV \cite{DVV} & 35.4 & 33.1 & 30.3 & 40.0 & 31.7 & 30.9 & 30.0 & 36.2 & 31.1 & 32.5 & 40.6 & 32.0 & 33.7 \\ \hline
CVP \cite{CVP} & 36.0 & 34.7 & 35.0 & 43.5 & 33.9 & 35.2 & 40.4 & 36.3 & 36.3 & 38.0 & 40.6 & 37.7 & 37.3 \\ \hline
HON4D \cite{HON4D} & 31.1 & 23.0 & 21.9 & 10.0 & 36.6 & 32.6 & 47.0 & 22.7 & 36.6 & 16.5 & 41.4 & 26.8 & 28.9 \\ \hline
SNV \cite{SNV} & 31.9 & 25.7 & 23.0 & 13.1 & 38.4 & 34.0 & 43.3 & 24.2 & 36.9 & 20.3 & 38.6 & 29.0 & 29.9 \\ \hline
HOPC \cite{HOPC} & 52.7 & 51.8 & 59.0 & 57.5 & 42.8 & 44.2 & 58.1 & 38.4 & 63.2 & 43.8 & 66.3 & 48.0 & 52.2 \\ \hline
HPM+TM \cite{HPM+TM} & 80.6 & 80.5 & 75.2 & 82.0 & 65.4 & 72.0 & 77.3 & 67.0 & 83.6 & 81.0 & 83.6 & 74.1 & 76.9 \\ \hline
\multicolumn{ 14}{|c|}{\textbf{Input: RGB + Depth images}} \\ \hline
Ours (RGB-D) & \textbf{86.9} & \textbf{89.8} & \textbf{81.9} & \textbf{89.5} & \textbf{76.7} & \textbf{83.6} & \textbf{83.6} & \textbf{79.0} & \textbf{89.6} & \textbf{82.1} & \textbf{89.2} & \textbf{83.8} & \textbf{84.6} \\ \hline
\end{tabular}}
\label{tab:uwa_comp_1}
\end{table}


\subsection{Northwestern-UCLA Dataset}
\begin{figure}[t]
\centering
\includegraphics[width=0.8\textwidth]{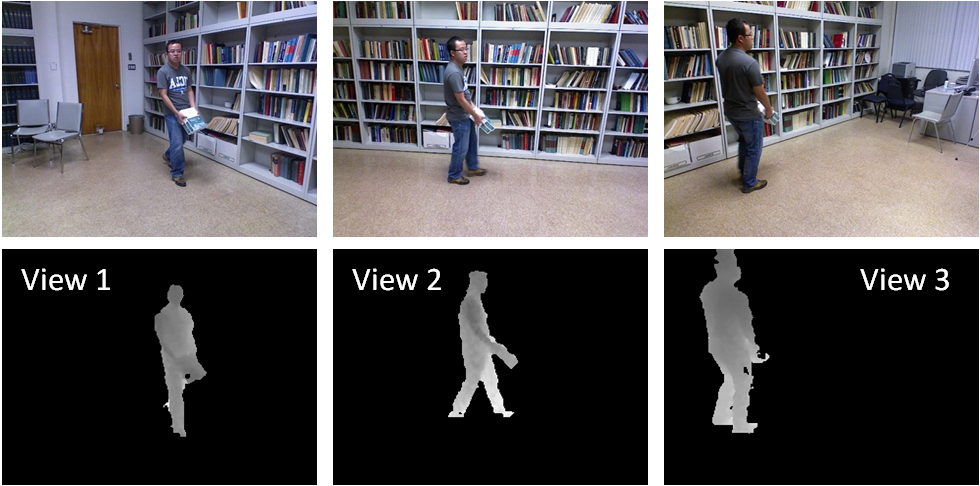}
\caption{RGB and Depth pairs of a single action from three different views in the Northwestern-UCLA Multiview Action 3D Dataset~\cite{AOG}.}
\label{fig:nucla_sample}
\end{figure}

The Northwestern-UCLA Multiview Action 3D Dataset~\cite{AOG} contains RGB-D videos captured simultaneously by 3 Kinect cameras from 3 different viewpoints. There are 10 action classes: (1) pick up with one hand, (2) pick up with two hands, (3) drop trash, (4) walk around, (5) sit down, (6) stand up, (7) donning, (8) doffing, (9) throw, and (10) carry. The three viewpoints available in the dataset are: (a) left, (b) front, and (c) right.
Every action in this dataset is performed by 10 subjects. In Fig.~\ref{fig:nucla_sample}, we show representative sample RGB and Depth image pairs from the three viewpoints. This dataset is  challenging for two reasons. Firstly, many action categories share the same ``walking'' pattern before and after the actual occurrence of the action of interest. Secondly, some actions such as ``pick up with on hand'' and ``pick up with two hands'' are very similar, which makes them to distinguished in the presence of the viewpoint variations.

For this dataset, we also use videos captured from two different views for training and the third view for testing. 
Again, we first provide the relative performance comparison of the proposed approach using the three network  architectures shown in Fig~\ref{fig:caffe_nktm_modes} in  
Table \ref{tab:nucla_comp_nktm}.
We can see that NKTM \#1 is also  the most suited  architecture for this dataset. 
In Table \ref{tab:nucla_comp_1}, we summarize  the performance comparison of the proposed method with the existing methods.
The proposed RGB-D method achieves an average accuracy of 82.7\%, which is 3.0\% higher than the nearest competitor HPM+TM.

\begin{table}[t]
\centering
\caption{Comparison of action recognition accuracy (\%) on the Northwestern-UCLA dataset using different NKTM networks from Figure \ref{fig:caffe_nktm_modes}.}
{\footnotesize
\begin{tabular}{|l|c|c|c|c|}
\hline
Training view & V1\&V2 & V1\&V3 & V2\&V3 & \multicolumn{ 1}{c|}{Mean} \\ \cline{ 1- 4}
Testing view & V3 & V2 & V1 & \multicolumn{ 1}{c|}{} \\ \hline
\multicolumn{ 5}{|c|}{\textbf{Input: RGB + Depth images}} \\ \hline
NKTM \#1 & {\bf 92.9} & {\bf 82.8} & {\bf 72.5} & {\bf 82.7} \\ \hline
NKTM \#2 & 92.5 & 82.4 & 71.9 & 82.3 \\ \hline
NKTM \#3 & 92.9 & 82.6 & 71.3 & 82.3 \\ \hline
\end{tabular}}
\label{tab:nucla_comp_nktm}
\end{table}

\begin{table}[t]
\centering
\caption{Comparison of action recognition accuracy (\%) on the Northwestern-UCLA Multiview Action 3D Dataset.}
{\footnotesize
\begin{tabular}{|l|c|c|c|c|}
\hline
Training view & V1\&V2 & V1\&V3 & V2\&V3 & \multicolumn{ 1}{c|}{Mean} \\ \cline{ 1- 4}
Testing view & V3 & V2 & V1 & \multicolumn{ 1}{c|}{} \\ \hline
\multicolumn{ 5}{|c|}{\textbf{Input: RGB images}} \\ \hline
Hankelets \cite{Hankelets} & 45.2 & - & - & - \\ \hline
DVV \cite{DVV} & 58.5 & 55.2 & 39.3 & 51.0 \\ \hline
CVP \cite{CVP} & 60.6 & 55.8 & 39.5 & 52.0 \\ \hline
AOG \cite{AOG} & 73.3 & - & - & - \\ \hline
nCTE \cite{nCTE} & 68.6 & 68.3 & 52.1 & 63.0 \\ \hline
NKTM \cite{NKTM} & 75.8 & 73.3 & 59.1 & 69.4 \\ \hline
\multicolumn{ 5}{|c|}{\textbf{Input: Depth images}} \\ \hline
DVV \cite{DVV} & 52.1 & - & - & - \\ \hline
CVP \cite{CVP} & 53.5 & - & - & - \\ \hline
HON4D \cite{HON4D} & 39.9 & - & - & - \\ \hline
SNV \cite{SNV} & 42.8 & - & - & - \\ \hline
HOPC \cite{HOPC} & 80.0 & - & - & - \\ \hline
HPM+TM \cite{HPM+TM} & 92.2 & 78.5 & 68.5 & 79.7 \\ \hline
\multicolumn{ 5}{|c|}{\textbf{Input: RGB + Depth images}} \\ \hline
Ours (RGB-D) & \textbf{92.9} & \textbf{82.8} & \textbf{72.5} & \textbf{82.7} \\ \hline
\end{tabular}}
\label{tab:nucla_comp_1}
\end{table}

\subsection{NTU RGB+D Human Activity Dataset}
\begin{figure}[t]
\centering
\includegraphics[width=0.8\textwidth]{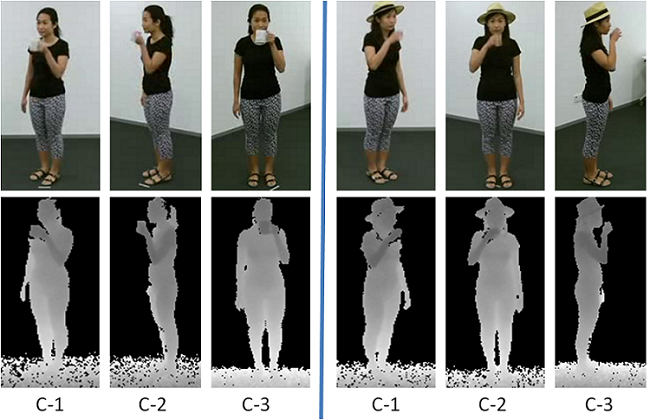}
\caption{RGB and depth samples  from the NTU RGB+D Human Activity Dataset \citep{shahroudy2016ntu}. Three sensors C-1, C-2 and C-3 are used to record the data. The left group of images shows the actions recorded with the performer facing the sensor C-3, and the right group of images are recorded when the action performer faces the sensor C-2.}
\label{fig:ntu_samples}
\end{figure}

The NTU RGB+D Human Activity Dataset \citep{shahroudy2016ntu} is a large-scale RGB+D dataset for human activity analysis. It was collected using the Kinect v2 sensor and it includes 56,880 action samples each for the RGB videos,  depth videos,  skeleton sequences and the infra-red videos. In our experiments, we only make use of the RGB and the depth videos from the dataset. In these videos, there are 40 human subjects performing 60 types of actions including 50 single person actions and 10 two-person interactions. Three sensors were used to capture the data simultaneously from three horizontal angles: $-45^\circ, 0^\circ, 45^\circ$, and every action performer performed the action twice, facing the left or the right sensor. 
Moreover, the height of the sensors and their distances to the action performers were also  adjusted to get further  variations in the viewpoints. The NTU RGB+D dataset is one of the largest and the most complex cross-view action dataset of its kind to date. Figure \ref{fig:ntu_samples} illustrates the RGB and the depth sample frames in the NTU RGB+D dataset.

We follow the standard evaluation protocol~ \citep{shahroudy2016ntu}, which includes cross-subject and cross-view evaluations. For the cross-subject protocol, 40 subjects are partitioned into the training and the testing groups, where each group consists of 20 subjects. For the cross-view protocol, the videos captured by the sensor C-2 and C-3 are used as the training samples, and the videos captured by the sensor C-1 are used as the testing samples.

Table \ref{tab:ntu_rgbd_comp} summarizes the comparison of our method with the existing approaches on the NTU dataset. 
The second column of the table indicates the data type used by the approaches from this dataset.
As can be seen, most of the existing approaches reporting result on this challenging dataset mainly exploit skeletal information.
Using only the RGB frames or the Depth frames generally does not result in reasonable performance on this dataset.
Similar to our approach, the DSSCA-SSLM~\cite{shahroudy2017deep} also uses both RGB and Depth information in their approach. 
However, the proposed method is able to achieve a significant improvement over the performance of that DSSCA-SSLM.


\begin{table}
\centering
{\caption{Action recognition accuracy (\%) on the NTU RGB+D Human Activity Dataset.}
\label{tab:ntu_rgbd_comp}}
{
\begin{tabular}{|l|l|c|c|}
\hline
Method & Data type & Cross Subject & Cross View \\ \hline
\multicolumn{ 3}{|c}{\textbf{Baseline}} &  \\ \hline

HON4D~\citep{HON4D} & Depth & 30.6 & 7.3 \\ \hline
SNV~\citep{yang2014super} & Depth & 31.8 & 13.6 \\ \hline
HOG-2~\citep{ohn2013joint} & Depth & 32.4 & 22.3 \\ \hline
Skeletal Quads~\citep{evangelidis2014skeletal} & Joints & 38.6 & 41.4 \\ \hline
Lie Group~\citep{vemulapalli2014human} & Joints & 50.1 & 52.8 \\ \hline
Deep RNN~\citep{shahroudy2016ntu} & Joints & 56.3 & 64.1 \\ \hline
HBRNN-L~\citep{du2015hierarchical} & Joints & 59.1 & 64.0 \\ \hline
Dynamic Skeletons~\citep{hu2015jointly} & Joints & 60.2 & 65.2 \\ \hline
Deep LSTM~\citep{shahroudy2016ntu} & Joints & 60.7 & 67.3 \\ \hline
LieNet~\citep{huang2016deep} & Joints & 61.4 & 67.0 \\
P-LSTM~\citep{shahroudy2016ntu} & Joints & 62.9 & 70.3 \\ \hline
LTMD~\citep{luo2017unsupervised} & Depth & 66.2 & - \\ \hline
ST-LSTM~\citep{liu2016spatio} & Joints & 69.2 & 77.7 \\ \hline
Two-stream RNN~\citep{wang2017modeling} & Joints & 71.3 & 79.5 \\ \hline
Res-TCN~\citep{kim2017interpretable} & Joints & 74.3 & \textbf{83.1} \\ \hline
DSSCA-SSLM~\citep{shahroudy2017deep} & RGB-D & \textbf{74.9} & - \\ \hline

\multicolumn{ 3}{|c}{\textbf{Proposed}} & \multicolumn{1}{l|}{} \\ \hline

Ours (RGB-D) & RGB-D & \bf{77.5} & \bf{84.5} \\ \hline

\end{tabular}
}
\end{table}

\section{Conclusion}
\label{sec:Conc}
We proposed an RGB-D human action recognition method that capitalizes on the view invariant characteristics of both Depth and RGB data streams, thereby making action recognition largely insensitive to the viewpoint variations in the videos.
The proposed method processes the RGB and Depth streams separately to fully exploit the individual modalities. We extract dense action trajectories are using the RGB frames to encode motion information, and then pass them through a deep network to get the viewpoint invariant features. For the Depth frames, we exploit the human pose model~\cite{HPM+TM} to extract the appearance information. The CNN fc$_7$ layer viewpoint invariant features are encoded by Fourier Temporal Pyramid to incorporate the temporal dimension. Spatio-temporal features from both RGB and Depth streams are normalized and combined to form a set of heterogeneous features that are used to collaboratively represent the test features. A convex combination of dense and sparse collaborative representations is eventually used to predict the label of the test feature. Experiments on three standard multi-view RGB-D dataset and comparison to twelve existing methods demonstrate the effectiveness of the proposed approach.

\section*{Acknowledgement}
This research was supported by ARC grant DP160101458.

\newpage 
\bibliography{jian_wacv_2017}

\begin{thebibliography}{10}
\expandafter\ifx\csname url\endcsname\relax
  \def\url#1{\texttt{#1}}\fi
\expandafter\ifx\csname urlprefix\endcsname\relax\def\urlprefix{URL }\fi
\expandafter\ifx\csname href\endcsname\relax
  \def\href#1#2{#2} \def\path#1{#1}\fi

\bibitem{nCTE}
A.~Gupta, J.~Martinez, J.~J. Little, R.~J. Woodham, 3d pose from motion for
  cross-view action recognition via nonlinear circulant temporal encoding, in:
  CVPR, 2014.

\bibitem{Dense_traj_1}
H.~Wang, A.~Klser, C.~Schmid, C.~Liu, Action recognition by dense trajectories,
  in: CVPR, 2011.

\bibitem{Dense_traj_2}
H.~Wang, A.~Klser, C.~Schmid, C.~Liu, Dense trajectories and motion boundary
  descriptors for action recognition, in: IJCV, 2013.

\bibitem{Dense_traj_3}
H.~Wang, C.~Schmid, Action recognition with improved trajectories, in: ICCV,
  2013.

\bibitem{HPM+TM}
H.~Rahmani, A.~Mian, 3d action recognition from novel viewpoints, in: CVPR,
  2016.

\bibitem{NKTM}
H.~Rahmani, A.~Mian, Learning a non-linear knowledge transfer model for
  cross-view action recognition, in: CVPR, 2015.

\bibitem{Chaaraoui_2013_ICCVW}
A.~Chaaraoui, J.~Padilla-Lopez, F.~Fl{\'o}rez-Revuelta, Fusion of skeletal and
  silhouette-based features for human action recognition with rgb-d devices,
  in: ICCVW, 2013.

\bibitem{Hu_2015_CVPR}
J.-F. Hu, W.-S. Zheng, J.~Lai, J.~Zhang, Jointly learning heterogeneous
  features for rgb-d activity recognition, in: CVPR, 2015.

\bibitem{Liu_2013_IJCAI}
L.~Liu, L.~Shao, Learning discriminative representations from rgb-d video data,
  in: IJCAI, 2013.

\bibitem{Shahroudy_2014_ISCCSP}
A.~Shahroudy, G.~Wang, T.-T. Ng, Multi-modal feature fusion for action
  recognition in rgb-d sequences, in: ISCCSP, 2014.

\bibitem{Koppula_2013_ICML}
H.~S. Koppula, A.~Saxena, Learning spatio-temporal structure from rgb-d videos
  for human activity detection and anticipation, in: ICML, 2013.

\bibitem{Wei_2013_ICCV}
P.~Wei, Y.~Zhao, N.~Zheng, S.-C. Zhu, Modeling 4d human-object interactions for
  event and object recognition, in: ICCV, 2013.

\bibitem{DICTA}
J.~Liu, N.~Akhtar, A.~Mian, View point invariant rgb-d human action
  recognition, in: International Conference on Digital Image Computing:
  Techniques and Applications (DICTA), 2017.

\bibitem{jia2014caffe}
Y.~Jia, E.~Shelhamer, J.~Donahue, S.~Karayev, J.~Long, R.~Girshick,
  S.~Guadarrama, T.~Darrell, Caffe: Convolutional architecture for fast feature
  embedding, arXiv preprint arXiv:1408.5093.

\bibitem{CRC}
L.~Zhang, M.~Yang, X.~Feng, Sparse representation or collaborative
  representation: Which helps face recognition?, in: International Conference
  on Computer Vision, 2011.

\bibitem{rw_3}
F.~Lv, R.~Nevatia, Single view human action recognition using key pose matching
  and viterbi path searching, in: CVPR, 2007.

\bibitem{rw_4}
T.~Syeda-Mahmood, A.~Vasilescu, S.~Sethi, Action recognition from arbitrary
  views using 3d exemplars, in: ICCV, 2007.

\bibitem{rw_5}
A.~Yilmaz, M.~Shah, Action sketch: a novel action representation, in: CVPR,
  2005.

\bibitem{Hankelets}
B.~Li, O.~Camps, M.~Sznaier, Cross-view activity recognition using hankelets,
  in: CVPR, 2012.

\bibitem{rw_7}
V.~Parameswaran, R.~Chellappa, View invariance for human action recognition,
  in: IJCV, 2006.

\bibitem{rw_8}
C.~Rao, A.~Yilmaz, M.~Shah, View-invariant representation and recognition of
  actions, in: IJCV, 2002.

\bibitem{rw_9}
D.~Weinland, R.~Ronfard, E.~Boyer, Free viewpoint action recognition using
  motion history volumes, in: CVIU, 2006.

\bibitem{rw_10}
A.~Farhadi, M.~K. Tabrizi, Learning to recognize activities from the wrong view
  point, in: ECCV, 2008.

\bibitem{rw_11}
A.~Farhadi, M.~K. Tabrizi, I.~Endres, D.~A. Forsyth, A latent model of
  discriminative aspect, in: ICCV, 2009.

\bibitem{rw_12}
R.~Gopalan, R.~Li, R.~Chellapa, Domain adaption for object recognition: An
  unsupervised approach, in: ICCV, 2011.

\bibitem{DVV}
R.~Li, T.~Zickler, Discriminative virtual views for crossview action
  recognition, in: CVPR, 2012.

\bibitem{rw_14}
J.~Liu, M.~Shah, B.~Kuipersy, S.~Savarese, Cross-view action recognition via
  view knowledge transfer, in: CVPR, 2011.

\bibitem{CVP}
Z.~Zhang, C.~Wang, B.~Xiao, W.~Zhou, S.~Liu, C.~Shi, Cross-view action
  recognition via a continuous virtual path, in: CVPR, 2013.

\bibitem{rw_16}
J.~Zheng, Z.~Jiang, Learning view-invariant sparse representations for
  cross-view action recognition, in: ICCV, 2013.

\bibitem{Wang_2015_CVPR}
L.~Wang, Y.~Qiao, X.~Tang, Action recognition with trajectory-pooled
  deep-convolutional descriptors, in: CVPR, 2015.

\bibitem{rw_19}
W.~Li, Z.~Zhang, Z.~Liu, Action recognition based on a bag of 3d points, in:
  CVPRW, 2010.

\bibitem{rw_17}
F.~Lv, R.~Nevatia, Recognition and segmentation of 3-d human action using hmm
  and multi-class adaboost, in: ECCV, 2006.

\bibitem{Lu_2014_CVPR}
C.~Lu, J.~Jia, C.-K. Tang, Range-sample depth feature for action recognition,
  in: CVPR, 2014.

\bibitem{HOPC}
H.~Rahmani, A.~Mahmood, D.~Q. Huynh, A.~Mian, Hopc: Histogram of oriented
  principal components of 3d pointclouds for action recognition, in: ECCV,
  2014.

\bibitem{SNV}
X.~Yang, Y.~Tian, Super normal vector for activity recognition using depth
  sequences, in: CVPR, 2014.

\bibitem{Kong_2015_CVPR}
Y.~Kong, Y.~Fu, Bilinear heterogeneous information machine for rgb-d action
  recognition, in: CVPR, 2015.

\bibitem{HPM_code}
Pretrained human pose model,
  \url{http://staffhome.ecm.uwa.edu.au/~00053650/code.html}.

\bibitem{AlexNet}
A.~Krizhevsky, I.~Sutskever, G.~E. Hinton, Imagenet classification with deep
  convolutional neural networks, in: NIPS, 2012.

\bibitem{CMU}
Cmu motion capture database, \url{http://http://mocap.cs.cmu.edu/}.

\bibitem{rw_21}
J.~Wang, Z.~Liu, Y.~Wu, J.~Yuan, Learning actionlet ensemble for 3d human
  action recognition, in: PAMI, 2013.

\bibitem{NA_PR}
N.~Akhtar, F.~Shafait, A.~Mian, Efficient classification with sparsity
  augmented collaborative representation, Pattern Recognition 65 (2017) 136 --
  145.

\bibitem{PAMI}
N.~Akhtar, F.~Shafait, A.~Mian, Discriminative bayesian dictionary learning for
  classification, IEEE Transactions on Pattern Analysis and Machine
  Intelligence 38~(12) (2016) 2374--2388.

\bibitem{NA_CVPR2017}
N.~Akhtar, A.~Mian, F.~Porikli, Joint discriminative bayesian dictionary and
  classifier learning, in: IEEE Int. Conf. on Computer Vision and Pattern
  Recognition (CVPR), 2017.

\bibitem{OMP}
J.~A. Tropp, A.~C. Gilbert, Signal recovery from random measurements via
  orthogonal matching pursuit, IEEE Transactions on Information Technology
  53~(12) (2007) 4655--4666.

\bibitem{HOPC2016PAMI}
H.~Rahmani, A.~Mahmood, D.~Huynh, A.~Mian, Histogram of oriented principal
  components for cross-view action recognition, in: PAMI, 2016.

\bibitem{AOG}
J.~Wang, X.~N. nad Y.~Xia, Y.~Wu, S.~Zhu, Cross-view action modeling, learning
  and recognition, in: CVPR, 2014.

\bibitem{shahroudy2016ntu}
A.~Shahroudy, J.~Liu, T.-T. Ng, G.~Wang, Ntu rgb+d: A large scale dataset for
  3d human activity analysis, in: IEEE Conference on Computer Vision and
  Pattern Recognition, 2016, pp. 1010--1019.

\bibitem{Action_Tube}
G.~Gkioxari, J.~Malik, Finding action tubes, in: CVPR, 2015.

\bibitem{LRCN}
J.~Donahue, L.~A. Hendricks, S.~Guadarrama, M.~Rohrbach, S.~Venugopalan,
  K.~Saenko, T.~Darrell, Long-term recurrent convolutional networks for visual
  recognition and description, in: CVPR, 2015.

\bibitem{HON4D}
O.~Oreifej, Z.~Liu, Hon4d: histogram of oriented 4d normals for activity
  recognition from depth sequences, in: CVPR, 2013.

\bibitem{shahroudy2017deep}
A.~Shahroudy, T.-T. Ng, Y.~Gong, G.~Wang, Deep multimodal feature analysis for
  action recognition in rgb+ d videos, IEEE Transactions on Pattern Analysis
  and Machine Intelligence.

\bibitem{yang2014super}
X.~Yang, Y.~Tian, Super normal vector for activity recognition using depth
  sequences, in: IEEE Conference on Computer Vision and Pattern Recognition,
  2014, pp. 804--811.

\bibitem{ohn2013joint}
E.~Ohn-Bar, M.~Trivedi, Joint angles similarities and hog2 for action
  recognition, in: IEEE Conference on Computer Vision and Pattern Recognition
  Workshops, 2013, pp. 465--470.

\bibitem{evangelidis2014skeletal}
G.~Evangelidis, G.~Singh, R.~Horaud, Skeletal quads: Human action recognition
  using joint quadruples, in: International Conference on Pattern Recognition,
  2014, pp. 4513--4518.

\bibitem{vemulapalli2014human}
R.~Vemulapalli, F.~Arrate, R.~Chellappa, Human action recognition by
  representing 3d skeletons as points in a lie group, in: IEEE Conference on
  Computer Vision and Pattern Recognition, 2014, pp. 588--595.

\bibitem{du2015hierarchical}
Y.~Du, W.~Wang, L.~Wang, Hierarchical recurrent neural network for skeleton
  based action recognition, in: IEEE Conference on Computer Vision and Pattern
  Recognition, 2015, pp. 1110--1118.

\bibitem{hu2015jointly}
J.-F. Hu, W.-S. Zheng, J.~Lai, J.~Zhang, Jointly learning heterogeneous
  features for rgb-d activity recognition, in: IEEE conference on Computer
  Vision and Pattern Recognition, 2015, pp. 5344--5352.

\bibitem{huang2016deep}
Z.~Huang, C.~Wan, T.~Probst, L.~Van~Gool, Deep learning on lie groups for
  skeleton-based action recognition, in: IEEE Conference on Computer Vision and
  Pattern Recognition, 2016.

\bibitem{luo2017unsupervised}
Z.~Luo, B.~Peng, D.-A. Huang, A.~Alahi, L.~Fei-Fei, Unsupervised learning of
  long-term motion dynamics for videos, in: IEEE Conference on Computer Vision
  and Pattern Recognition, 2017.

\bibitem{liu2016spatio}
J.~Liu, A.~Shahroudy, D.~Xu, G.~Wang, Spatio-temporal lstm with trust gates for
  3d human action recognition, in: European Conference on Computer Vision,
  2016, pp. 816--833.

\bibitem{wang2017modeling}
H.~Wang, L.~Wang, Modeling temporal dynamics and spatial configurations of
  actions using two-stream recurrent neural networks, in: IEEE Conference on
  Computer Vision and Pattern Recognition, 2017.

\bibitem{kim2017interpretable}
T.~S. Kim, A.~Reiter, Interpretable 3d human action analysis with temporal
  convolutional networks, in: IEEE Conference on Computer Vision and Pattern
  Recognition Workshop, 2017.

\end{thebibliography}

\end{document}